\documentclass[a4paper,twoside]{article}

\usepackage{epsfig}
\usepackage{array}
\usepackage{subcaption}
\usepackage{calc}
\usepackage{cite}
\usepackage{amssymb}
\usepackage{amstext}
\usepackage{amsmath}
\usepackage{amsthm}
\usepackage{multicol}
\usepackage{pslatex}
\usepackage{apalike}
\usepackage[bottom]{footmisc}
\usepackage{SCITEPRESS}     

\begin{document}

\title{Challenges in Domain-Specific Abstractive Summarization and How to Overcome them}
\author{\authorname{Anum Afzal\sup{1}, Juraj Vladika\sup{1}, Daniel Braun\sup{2}, Florian Matthes\sup{1}}
\affiliation{\sup{1}Department of Computer Science, Technical University of Munich, Boltzmannstrasse 3, 85748 Garching bei Muenchen, Germany}
\affiliation{\sup{2}Department of High-tech Business and Entrepreneurship, University of Twente, Hallenweg 17, 7522NH  Enschede, The Netherlands}
\email{\{anum.afzal, juraj.vladika, matthes\}@tum.de, d.braun@utwente.nl}
}

\keywords{Text Summarization, Natural Language Processing, Efficient Transformers, Model Hallucination, Natural Language Generation Evaluation, Domain-adaptation of Language Models}

\abstract{Large Language Models work quite well with general-purpose data and many tasks in Natural Language Processing. However, they show several limitations when used for a task such as domain-specific abstractive text summarization. This paper identifies three of those limitations as research problems in the context of abstractive text summarization: 1) Quadratic complexity of transformer-based models with respect to the input text length; 2) Model Hallucination, which is a model's ability to generate factually incorrect text; and 3) Domain Shift, which happens when the distribution of the model's training and test corpus is not the same. Along with a discussion of the open research questions, this paper also provides an assessment of existing state-of-the-art techniques relevant to domain-specific text summarization to address the research gaps.}

\onecolumn \maketitle \normalsize \setcounter{footnote}{0} \vfill

\section{\uppercase{Introduction}}
\label{sec:introduction}

With the ever-increasing amount of textual data being created, stored, and digitized, companies and researchers have large corpora at their disposal that could be processed into useful information. Perusal and encapsulation of such data usually require domain expertise which is costly and time-consuming. Abstractive text summarization using Natural Language Processing (NLP) techniques, is a powerful tool that can provide aid for this task. Unlike the traditional automatic text summarization techniques, which extracts the most relevant sentences from the original document, abstractive text summarization generates new text as summaries. For the sake of simplicity, the term text summarization would be used to represent abstractive text summarization in this paper.


While text summarization \cite{abssummarization-survey,Klymenko-2020-Text-Summarization-Review} on general textual data has been an active research field in the past decade, summarization of domain-specific documents, especially to support business and scientific processes have not received much attention.
State-of-the-art research focuses on deep learning models in NLP to capture semantics and context associated with the text. While these Large Language Models (LLMs) perform well on the general-purpose corpus, their performance declines when tested against domain-specific corpus. 
This paper discusses some challenges LLMs face in the context of a text summarization task and provides an overview of existing techniques that could be leveraged to counter those challenges.

Previous research in text summarization has mostly focused on general-purpose data \cite{abssummarization-survey,Allahyari-Text-Summarization-Survey}. Domain-specific summarization however, is still an active research area and has many research questions that need to be addressed. This paper addresses some of those theoretical research questions and provides an initial assessment of the existing techniques can be utilized to overcome those limitations. We have identified three important hurdles with regards to the successful implementation of an NLP model for generation of domain-specific summaries. 
\begin{enumerate}

    \item Most language models have quadratic complexity, meaning that the memory requirement grows quadratically as the length of the text increases. As a result, transformer-based language models are not capable of processing large documents. Since most documents that need to be summarized are long, it creates a need for language models capable of handling them efficiently without overshooting in terms of complexity. 
        
    \item Evaluating generated summaries is difficult using common evaluation metrics, that look at word overlaps between generated summaries and the reference text. This curbs model expressiveness in favor of repeating the original human wording. Generated summaries can include information not present in the original document, a phenomenon known as model hallucination. Factually incorrect summaries are problematic in domains like science or journalism because they can produce misinformation and reduce the trust in models. 
 
    \item State-of-the-art text summarization models are pre-trained on general-purpose corpus and hence do not perform well on domain-specific text. This happens because a domain-specific text contains words and concepts that were not a part of the original model training vocabulary. When generating summaries, it is essential for the model to encode the text properly, which is usually not the case since the model fails to capture domain-specific concepts.

\end{enumerate}

\noindent Hence, to produce concise and meaningful domain-specific summaries, it is important to address the following three research gaps:

\begin{itemize}
  \item How to overcome the input size limitation of transformer-based language model so they can process large documents without running into complexity issues?
  \item How to evaluate summaries generated by a language model to ensure they convey all the important information while being factually correct?
  \item  How can we adapt an existing general-purpose language model to understand the underlying concepts and vocabulary of the new domain?
\end{itemize}

\noindent This paper is divided into five sections. The first section provided an introduction to the topic and outlined three important hurdles faced in domain-specific summarization. Section \ref{section 2} builds up on the research gaps and further elaborates them. Section \ref{section 3} outlines the existing techniques that can be used to overcome those research gaps, followed by Section \ref{section 4} that initiates a comparative discussion on the existing techniques. Finally, Section \ref{section 5} concludes this paper and provides hints related to the future work.

\section{\uppercase{Current Challenges in Text Summarization}}
\label{section 2}

For a task such as text summarization, a sequence-to-sequence (Seq2Seq) architecture that takes text as input and produces text as output, is the most suitable one. Since traditional seq2seq models like Recurrent Neural Networks (RNNs) 
and Long short-term memory (LSTMs) \cite{lstm} have some inherent limitations, such as not being able to encode long-term dependencies in text and lack of parallelism opportunities, they are not suitable for domain-specific summarization of long documents. Transformer-based seq2seq models address these limitations by allowing computations to be parallelized, retain long-term dependencies via a self-attention matrix, and better text encoding through a word embedding module that has been trained on a huge corpus. 
As discussed in the section below, these models come with their own set of impediments when utilized for summarization of domain-specific long documents. 



\subsection{Transformers and their quadratic complexity}

First introduced in the paper \textit{Attention is all you need} \cite{attention}, the Transformers immediately became popular and have since been the backbone of LLMs. 
By design, they are pre-trained on a huge corpus allowing them to learn the patterns, context, and other linguistic aspects of the text. Furthermore, they are trained using self-supervised approaches that allow them to make use of the huge corpora of unstructured and unlabeled data. The heart of a Transformer block however, is the self-attention matrix that helps it retain the long-term context of the text. The self-attention matrix essentially tells the model how much attention a word should pay to all the other words in the text. While this information is vital, its calculation consumes a huge amount of memory and takes a long time to compute. The calculation of the $n \times n$ self-attention matrix, where $n$ is the number of tokens (sequence length), entails quadratic complexity.
As a workaround, the input text is typically truncated to retain only the first 512 tokens. For tasks such as text summarization, it is important for the model to encodes the entire input text and hence, this problem is still an open research area.

\subsection{NLG Evaluation and Hallucinations}
A common challenge in generating summaries from scratch is how to meaningfully evaluate their content and ensure factual consistency with the source text.

\subsubsection{Evaluating Summarization}
Natural Language Generation (NLG) is a subset of NLP dealing with tasks where new text is generated, one of them being abstractive summarization. The output of models for NLG tasks is notoriously hard to evaluate because there is usually a trade-off between the expressiveness of the model and its factual accuracy \cite{sai2022survey}. Metrics to evaluate generated text can be word-based, character-based, or embedding-based. Word-based metrics are the most popular evaluation metrics, owing to their ease of use. They look at the exact overlap of n-grams (n consecutive words) between generated and reference text. Their main drawback is that they do not take into account the meaning of the text. Two sentences such as “\textit{Berlin is the capital of Germany}” and “\textit{Berlin is not the capital of Germany}” have an almost complete n-gram overlap despite having opposite meanings.

\subsubsection{Model Hallucinations}

Even though modern transformer models can generate text that is coherent and grammatically correct, they are prone to generating content not backed by the source document. Borrowing the terminology from psychology, this is called model hallucination. In abstractive summarization, the summary is said to be hallucinated if it has any spans not supported by content in the input document \cite{maynez-etal-2020-faithfulness}. Two main types of hallucinations are (1) intrinsic, where the generated content contradicts the source document; and (2) extrinsic, which are facts that cannot be verified from the source document. For example, if the document mentions an earthquake that happened in 2020, an intrinsic hallucination is saying it happened in 2015, while an extrinsic one would be a sentence about a flood that is never even mentioned in the document. In their analysis of three recent state-of-the-art abstractive summarization systems, \cite{falke-etal-2019-ranking} show that 25\% of generated summaries contain hallucinated content. Hallucinations usually stem from pre-trained large models introducing facts they learned during their training process, which is unrelated to a given source document.

 \subsection{Domain Shift in Natural Language Processing }

When working with specific NLP applications, domain knowledge is paramount for success. Finding labeled training data, or even unlabeled data in some cases, is a big challenge. Training data is often scarce in many domains/languages and often hinders the solution development for domain-specific tasks in NLP. Transfer Learning 
provides a solution to this by utilizing the existing model knowledge and building on it when training the model for a new task. Essentially, it allows the transfer and adaptation of the knowledge acquired from one set of domains and tasks to another set of domains and tasks.

Transformer-based language models in tandem with Transfer Learning have proven to be quite successful in the past years and have found their application in several real-world use cases. While they work well with tasks involving general-purpose corpus, there is a performance decline when it comes to domain-specific data. This happens because these language models are pre-trained on general-purpose data but are then tested on a domain-specific corpus. This difference in the distribution of training and testing data is known as the Domain Shift problem in NLP. It essentially means that the model doesn't know the domain-specific corpus contains words and concepts since they were not part of model's pre-training. 

\section{\uppercase{Existing Techniques }}
\label{section 3}

This section presents an overview of the existing techniques and architectures that can be applied for the summarization of domain-specific documents. These techniques are categorized into three sections based on the research questions they address; Efficient Transformers, Evaluation metrics, and Domain adaptation of Language Models. These techniques are summarized in Table \ref{tab:overview}, and discussed in detail in the section below.

\begin{table*}[ht]

\vspace{-0.2cm}
\caption{An overview of the research gaps, the proposed solutions, and the existing techniques that can be utilized for domain-specific abstractive summarization as discussed in Sections 2 and 3.}\label{tab:overview} \centering

\begin{tabular}{|>{\centering\arraybackslash} p{45mm}|>{\centering\arraybackslash} p{45mm}|>{\centering\arraybackslash} p{50mm}|}
  \hline
  \textbf{Challenges} & \textbf{Proposed Solution} & \textbf{Existing Techniques} \\
  \hline
   \begin{tabular}{@{}c@{}} Quadratic Complexity of \\ Transformer Models \end{tabular} & Efficient Transformers & \begin{tabular}{@{}c@{}c@{}c@{}} BigBird \\ Longformer Encoder-Decoder \\ Reformer, Performers \end{tabular}  \\
  \hline
   \begin{tabular}{@{}c@{}} NLG Evaluation and \\ Hallucination Mitigation \end{tabular}  &  \begin{tabular}{@{}c@{}} Semantic Evaluation Metrics \\ Fact-Checking\end{tabular} & \begin{tabular}{@{}c@{}}METEOR, BERTScore \\ NLI-based, QA-based \\ \end{tabular} \\
  \hline
    \begin{tabular}{@{}c@{}} Domain shift in \\ Language Models  \end{tabular} & \begin{tabular}{@{}c@{}} Domain-adaptation of \\ Language Models  \end{tabular} & \begin{tabular}{@{}c@{}c@{}} Fine-tuning-based \\ Pre-training-based \\ Tokenization-based \end{tabular} \\
  \hline
  
\end{tabular}
\end{table*}




\subsection{Efficient Transformers}

The quadratic complexity of the Transformer block is a well-known issue and several approaches to counter this have been proposed in the past years. All of these approaches focusing on adapting the self-attention mechanism of the Transformer block to reduce the quadratic complexity are categorized as Efficient Transformers. The survey by Tay et al. provides a detailed taxonomy of all available Efficient Transformers \cite{efficienttransformer-survey}. Some state-of-the-art Efficient Transformers suitable for domain-specific text summarization are discussed below:
\smallskip

\noindent {\bf BigBird.} 
BigBird is a long sequence Transformer that was introduced by Zaheer et al. and can process up to 4,096 tokens at a time. The attention mechanism of BigBird essentially consists of three parts in which all tokens attend to 1) a set of global tokens, 2) a set of randomly chosen tokens, and 3) all tokens in direct adjacency \cite{bigbird}. The set of global tokens attending to the entire sequence consists of artificially introduced tokens. The local attention is implemented in form of a sliding window of a predefined width $w$, in which a token attends to the $w/2$ preceding and following tokens in the sequence. The BigBird model's memory complexity is linear with regard to the length of the input sequence, i.e., it is $O(N)$ \cite{efficienttransformer-survey}.
\smallskip

\noindent {\bf Longformer Encoder-Decoder.}
The Longformer Encoder-Decoder (LED) model is a variant of the Longformer for sequence-to-sequence tasks such as summarization or translation \cite{beltagy2020longformer}. Similar to the BigBird model, the original Longformer relies on a sliding window attention of width $w$ with each token attending to the $w/2$ preceding and following tokens in the sequence. Stacking multiple layers, each using sliding window attention, ensures that a large amount of contextual information is embedded in each token's encoding. Apart from sliding window attention, the authors also use dilated sliding window attention. This in effect reduces the resolution of the sequence and allows the model to include more contextual information with fixed computational costs. 
The Longformer model also incorporates global attention. Similar to BigBird's global attention, a set of predefined positions in the input sequence attend to the entire sequence and all tokens in the sequence attend to the same global tokens. LED has an encoder that uses the local+global attention pattern of the original Longformer and a decoder that uses the full self-attention on the encoding provided by the encoder. The LED model scales linearly as the input length increases and hence has a complexity of $O(N)$ \cite{efficienttransformer-survey}.
\smallskip

\noindent {\bf Reformer}
The Reformer \cite{kitaev2020reformer} follows a two-step approach to reduce the complexity of the Transformer block. Firstly, the Reformer model makes use of reversible residual networks RevNet \cite{gomez-2017-revnet} which allow the model to store only one instance of the activations rather than having to store activations for every layer to be able to use back-propagation. In RevNets any layer's activations can be restored from the ones of the following layer and the model's parameters \cite{gomez-2017-revnet} hence reducing the model's memory requirements drastically. Secondly, to reduce the quadratic complexity with regard to the input sequence's length, the authors use locality-sensitive hashing to approximate the attention matrix. The attention mechanism's outsized memory requirements result from the computation of the attention matrix, i.e., $softmax(\frac{QK^{T}} {\sqrt{d_k}})$, and in that mainly the computation of $QK^{T}$. The authors point out that applying the softmax function implies that the attention matrix is dominated by the largest elements of $QK^{T}$. These largest elements result from the dot-product of the query and key vectors that are most similar to each other. Kitaev et al. note that the attention matrix can, consequently, be efficiently approximated by only computing the dot-product of those query and key vectors with the closest distance to each other. The Reformer uses locality-sensitive hashing to determine the closest neighbors of each query vector. 
The memory complexity of the LSH attention mechanism is $O(N\log{}N)$ in the length of the input sequence \cite{efficienttransformer-survey}. 
\smallskip

\noindent {\bf Performers}. The Performer architecture relies on a mechanism known as \textit{Fast Attention Via positive Orthogonal Random features} (FAVOR+) to approximate the self-attention matrix in kernel space. This technique is different from the previously discussed ones since it does not make any assumptions about the behavior of the self-attention matrix such as low-rankness or sparsity and guarantees low estimation variance, uniform convergence, and an almost-unbiased estimation of the original self-attention matrix. 
The authors further state that the Performer is compatible with existing pre-trained language models and requires little further fine-tuning \cite{choromanski2020rethinking}. 
\noindent The Performer's complexity is $O(N)$ \cite{efficienttransformer-survey} in terms of time and space.

\subsection{Semantic Evaluation Metrics and Fact-Checking of Hallucinations}

Numerous metrics have been devised for evaluating generated summaries. Word-based metrics look at n-gram overlaps between a candidate summary and the source document, while semantic evaluation metrics take into account the meaning of generated words and sentences. Many state-of-the-art generative models for summarization produce hallucinations,
so there is an increasing effort to detect and eliminate them.

\subsubsection{Evaluation Metrics}
\textbf{Word-based metrics.} These metrics look at exact overlap between words in candidate summaries and gold summary. BLEU is a metric based on precision which computes the n-gram overlap between the generated and the reference text \cite{bleu2002}. It is calculated for different values of \textit{n} and for all generated candidate summaries that are to be evaluated. The final BLEU-N score is the geometric mean of all intermediate scores for all values of \textit{n}. ROUGE is a metric similar to BLEU, but it is based on recall instead of precision \cite{lin-2004-rouge}. This means that for any given \textit{n}, it counts the total number of n-grams across all the reference summaries, and finds out how many of them are present in the candidate summary.\smallskip

\noindent \textbf{Semantic evaluation metrics.} Since both BLEU and ROUGE look at exact word matching, this leaves no room for synonyms or paraphrases. METEOR is a metric \cite{banerjee-lavie-2005-meteor} that builds up on BLEU by relaxing the matching criteria. It takes into account word stems and synonyms, so that two n-grams can be matched even if they are not exactly the same. Moving away from synonym matching, embedding-based metrics capture the semantic similarity by using dense word/sentence embeddings, together with vector-based similarity measures (like cosine similarity), to evaluate how closely the summary matches the source text. BERTScore is one such metric that utilizes BERT-based contextual embeddings of generated text and reference text in order to calculate the similarity between them \cite{BERTscore}.

\subsubsection{Hallucination Detection}
Detecting hallucinations in generated summaries is still a challenging task, for which dedicated methods are developed.
Based on the availability of annotated training data, these approaches can be split into unsupervised and semi-supervised \cite{huang2021}.\smallskip

\noindent \textbf{Unsupervised metrics}. These metrics rely on repurposing approaches for other NLP tasks like information extraction (IE), natural language inference (NLI), or question answering (QA) for the task of hallucination detection. The motivation behind this is the availability of training datasets for these tasks as opposed to scarce datasets for hallucination detection. The IE-based metrics compare the sets of extracted triples (subject, relation, object) and named entities from both the source document and generated summary to detect hallucination \cite{goodrich2019}. The NLI-based approaches in try to determine whether the generated summary logically entails the source document with a high probability \cite{falke-etal-2019-ranking}. The QA-based approaches work by posing the same set of questions to both the original document and the generated summary, and then comparing the two sets of obtained answers. Intuitively, a non-hallucinated summary and the source document will provide similar answers to the posed questions \cite{gabriel-etal-2021-go}.\smallskip

\noindent \textbf{Semi-supervised metrics.} This type of metric relies on datasets designed specifically for the task of hallucination detection. The data is usually synthetically generated from existing summarization datasets. For example, the weakly-supervised model FactCC \cite{kryscinski-etal-2020-evaluating} was trained jointly on three tasks: sentence factual consistency, supporting evidence extraction from source, and incorrect span detection in generated summaries. Similarly, in \cite{zhou2021detecting} a transformer model was trained on synthetic data with inserted hallucinations with the task of predicting hallucinated spans in summaries.\smallskip

\subsubsection{Hallucination Mitigation}
The approaches to mitigate hallucinations in summarization can generally be divided into pre-processing methods, that try to modify the model architecture or training process so that models can generate more factually-aware summaries in the first place, and post-processing methods, that aim to correct hallucinations in already generated candidate summaries.\smallskip

\noindent \textbf{Pre-processing methods.} The main line of work here focuses on augmenting the model architecture by modifying the encoder or decoder component of sequence-to-sequence models. One way of enhancing the encoders is injecting external knowledge into them before the training process, such as world knowledge triples from Wikidata \cite{Gunel2020MindTF}.
For improving the decoding process, tree-based decoders were used \cite{Song2020JointPA}. Another line of research involves modifying the training process. For example, contrastive learning was used in \cite{cao-wang-2021-cliff}, where positive examples were human-written summaries and negative examples were hallucinatory, generated summaries.\smallskip

\noindent \textbf{Post-processing methods.} These methods approach the problem by detecting the incorrect facts in the first version of a generated summary and then correcting them for the final version. For this purpose, in \cite{chen-etal-2021-improving} contrast candidate generation was used to replace incorrect named entities in summaries with those entities present in the source document. 
One promising research direction that has not been explored a lot is applying methods of fact-checking for hallucination detection and correction. Such an approach was used in \cite{dziri-etal-2021-neural}, where responses of conversational agents were checked and factually corrected before being sent out to users. The task of automated fact-checking consists of assessing the veracity of factual claims based on evidence from external knowledge sources \cite{zeng2021}. It is usually performed in a pipeline fashion, where first relevant documents and sentences are retrieved as evidence, and then veracity is predicted by inferring if there is entailment between the claim and evidence. Recently, there is an increasing interest in automatically verifying claims related to science, medicine, and public health \cite{kotonya-toni-2020-explainable-automated}.

\subsection{Domain Adaptation of Language Models}
\label{domain-adaptation of LLM}

Domain adaptation of Language Models has been a hot research area recently giving rise to several approaches. Some approaches relevant to abstractive text summarization are discussed below:
\smallskip

\noindent \textbf{Fine-tuning-based.} The most commonly used approach involves fine-tuning a pre-trained language model on a smaller task-specific dataset. 
In general, fine-tuning means retraining an existing model to adjust its weights to the specific-domain dataset or task so the model can make better predictions. One such approach is portrayed by Karouzos et al. where they first employ continued training on a BERT architecture utilizing a Masked Language Model loss. This approach is different from standard fine-tuning approaches because it makes use of an unlabeled corpus for domain adaptation. As a second step, they fine-tune the domain-adapted model from the previous step on a classification task \cite{karouzos-2021-udalm}.
\smallskip

\noindent \textbf{Pre-training-based.} A pre-training-based approach as compared to a fine-tuning-based approach trains the model weights from scratch instead of continued training on previously trained weights. In the past years, there have been many research contributions in the area of text summarization but it has been mostly restricted to general-purpose corpus. One similar approach involving a pre-training-based approach is presented by the authors Moradi et al. where they utilize a combination of graph-based and embedding-based approaches for the extractive summarization of biomedical article \cite{MORADI-biomedical-domain-text-summarization}. To counter the domain shift problem, they first re-train a BERT architecture on medical documents to ensure the availability of domain-specific word embedding. Then they generate sentence-level embedding of the input documents using the previously re-trained model. To generate summaries, they employ a graph-based approach to assign weights to previously generated sentence-level embedding and followed a sentence ranking algorithm to select the candidates for the summary generation. 
Another similar approach related to multi-domain adaptive models is presented by Zhong et al. for a text summarization task. They use the existing BART\cite{bart} architecture and exploit the multitask learning objective (including text summarization, text reconstruction, and text classification) to expand the coverage area of the existing model without changing its architecture \cite{zhong-2022-mtl-das}.
\smallskip

\noindent \textbf{Tokenization-based.} A tokenization-based approach involves updating the model tokenizer \cite{song-2021-wordpiece,kudo-2018-sentencepiece} to either include new domain-specific words or influencing its algorithm to prioritize a sequence of sub-words belonging to the domain-specific corpus. While fine-tuning and pre-training is a basic yet powerful technique for domain adaptation, over the years, some authors have contributed to this problem by employing tokenization-based techniques. Sachidananda et al. for instance propose an alternate approach where they adapt the RoBERTa \cite{roberta} tokenizer to include words from the domain-specific corpus. Most tokenization schemes typically merge subwords to create an individual token if that combination has a higher frequency in the domain-specific corpus. Sachidananda et al. approach this by influencing the tokenizer to prioritize such subword sequences from the domain-specific corpus rather than the base corpus \cite{sachidananda-2021-adaptive-tokenization}.
\section{\uppercase{Discussion}}
\label{section 4}

While the end goal of all Efficient Transformers is to reduce the quadratic complexity of the self-attention matrix, the techniques employed by them can be categorized into 1) techniques that assume sparsity of the self-attention matrix and compute only a few relevant entries, or 2) techniques that take advantage of mathematical compositions of the self-attention matrix such as Low Rankness, transformation to a Kernel Space, and other optimizations to reduce the complexity. In general, all efficient transformers have performance close to the original transformer on benchmark datasets but their performance in the real-life application is yet to be evaluated.

\noindent Effectively evaluating generated summaries is an ongoing challenge. Recent embedding-based metrics such as BERTScore take into account the context and semantics of sentences and are better correlated with human judgment. Still, these metrics are way more computationally intensive, their score is dependent on the PLM used, and they lack the intuitive explainability that standard scores like BLEU or ROGUE provide. There are domains, such as legislative, where specific terms and sentence structure is important to be preserved in the summary, therefore classic word-based metrics are preferred for evaluating them.

\noindent To overcome the domain shift in LLMs, several techniques have been proposed by researchers. 
When working with LLMs, the availability of task-specific training data is a challenge. In most cases, the decision between fine-tuning or pre-training can be based on the availability of the training resources and data. If enough domain-specific training data and computing resources are available, pre-training a domain-specific model might always be the best choice. A tokenization-based approach can be used exclusively with a fine-tuning-based approach as an additional add-on to enhance performance.

\section{\uppercase{Conclusion and Future Work}}
\label{section 5}

We assume that domain-specific text summarization will gain importance in the research field of NLP due to its ability to automate the task of manual summarization. This paper is meant to serve as a foundation step for research along the three research gaps addressed. While there are several promising NLP models for abstractive text summarization \cite{pegasud,bart}, they are not efficient in their training techniques as the size of the input documents increases. Moreover, when tested on the domain-specific corpus, they suffer from the domain-shift problem and often hallucinate because they were trained on general-purpose corpora and lack domain knowledge. On top of that, the automatic evaluation of the generated text is still a challenge. To the best of our knowledge, there have been several contributions to each of these individual research gaps however, an integrated approach addressing them from a text summarization perspective is lacking. A domain-adapted efficient transformer architecture in tandem with external fact-checking mechanisms and better automatic evaluation metrics could drastically improve the performance of text summarization models. The future work could be contributions towards the individual research gaps with the end goal of an integrated solution for text summarization.



\bibliographystyle{apalike}
{\small
\bibliography{main}}



\end{document}